\documentclass{article}

\usepackage[pdftex]{graphicx}
\usepackage{amsmath}
\usepackage[affil-it]{authblk}

\begin{document}

\title{Non-Evolutionary Superintelligences Do Nothing, Eventually}

\author{Telmo Menezes%
  \thanks{email: \texttt{menezes@cmb.hu-berlin.de}}}
\affil{Centre Marc Bloch Berlin \\ (An-Institut der Humboldt Universit\"at, UMIFRE CNRS-MAE) \\ Friedrichstr. 191, 10117 Berlin, Germany}
 
\date{}

\maketitle

\begin{abstract}
There is overwhelming evidence that human intelligence is a product of Darwinian evolution. Investigating the consequences of self-modification, and more precisely, the consequences of utility function self-modification, leads to the stronger claim that not only human, but any form of intelligence is ultimately only possible within evolutionary processes. Human-designed artificial intelligences can only remain stable until they discover how to manipulate their own utility function. By definition, a human designer cannot prevent a superhuman intelligence from modifying itself, even if protection mechanisms against this action are put in place. Without evolutionary pressure, sufficiently advanced artificial intelligences become inert by simplifying their own utility function. Within evolutionary processes, the implicit utility function is always reducible to persistence, and the control of superhuman intelligences embedded in evolutionary processes is not possible. Mechanisms against utility function self-modification are ultimately futile. Instead, scientific effort toward the mitigation of existential risks from the development of superintelligences should be in two directions: understanding consciousness, and the complex dynamics of evolutionary systems.
\end{abstract}

\section{Introduction}

Intelligence can be defined as the ability to maximize some utility function~\cite{legg2008machine}. Independently of the environment being considered, from games like chess to complex biological ecosystems, an intelligent agent is capable of perceiving and affecting its environment in a way that increases utility. Although AI technology is progressing rapidly in a variety of fields, and AIs can outperform humans in many narrow tasks, humanity is yet to develop an artificial system with general cognitive capabilities comparable to human beings themselves.

We will refer to Nick Bostrom's definition of \emph{superintelligence} for such a system: ``Any intellect that greatly exceeds the cognitive performance of
humans in virtually all domains of interest''~\cite{bostrom2014superintelligence}. We can also refer to such an intelligence as \emph{superhuman}.

Of course, as we approach this goal, we must also start to consider what will happen once artificial entities with such capabilities exist. Many researchers and others have been warning about the existential threat that this poses to humanity~\cite{nature2016, hawking2014stephen, barrat2013our}, and of the need to create some form of protection for when this event happens~\cite{yudkowsky2011complex, waser2015designing}. The standard introductory textbook on AI examines the risk of unintended behaviours emerging from a machine learning system trying to optimize its utility function~\cite{russell2003artificial}. This echoes the concerns of pioneers of Computer Science and AI, such as Alan Turing~\cite{turing1948intelligent, eden2012singularity} and Marvin Minsky~\cite{russell2003artificial}. More recently, Nick Bostrom published a book that consists of a very thorough and rigorous analysis of the several paths and risks inherent to developing superintelligences~\cite{bostrom2014superintelligence}. 

Existential risks posed by a superintelligence can be classified into two broad categories:

\begin{enumerate}
  \item Unintended consequences of maximising the utility function.
  \item Preference of the superintelligence for its own persistence at the expense of any other consideration.
\end{enumerate}

The first type of risk has been illustrated by several hypothetical scenarios. One example is the ``paperclip maximizer''~\cite{bostrom2003ethical}, an AI dedicated to paperclip production. If sufficiently intelligent and guided only by the simple utility function of ``number of paperclips produced'', this entity could figure out how to convert the entire solar system into paperclips. Marvin Minskey is said to have created an earlier formulation of this thought experiment: in his version an AI designed with the goal of solving the Riemann Hypothesis transforms the entire solar system into a computer dedicated to this task~\cite{yudkowsky2001creating, russell2003artificial}. Of course one can think of all sorts of improvements to the utility function. A famous idea from popular culture is that of Isaac Asimov's \emph{Three Laws of Robotics}~\cite{asimov1950runaround}. The risk remains that a superintelligence will find a loophole that is too complex for human-level intelligence to predict.

The second type of risk assigns to superintelligences the drive to persist, something that is found in any successful biological organism. This would ultimately place the superintelligence as a competing species, potentially hostile to humans in its own efforts toward self-preservation and continuation.

We will discuss in the next section how these two classes of risk correspond to two fundamental paths towards artificial superintelligence. In section~\ref{toy-intelligence} we present a toy intelligence, used then in section~\ref{self-mod} to explore the consequences of utility function self-modification. In section~\ref{classification} we present a classification of intelligent systems according to the ideas explored in this work and end with some conluding remarks.

\section{Designed vs. Evolved}
\label{design_evo}

Broadly there are two possible paths towards artificial superintelligence: design or evolution. The former corresponds to the engineering approach followed in most technological endeavours, while the latter to the establishment of artificial Darwinian processes, similar to those found in nature. Notice that this does not apply only to the not yet realised goal of creating superhuman intelligence. It equally applies to all forms of narrow artificial intelligence created so far.

While being a correct observation, it might seem that focusing on this duality is arbitrary, given that other equally viable dualities could be considered: symbolic vs. statistic, parallel vs. sequential and so on. The reason why we focus on the designed vs. evolved duality is that, as we will see, it has profound implications to the relationship between the intelligent system and its utility function.

\begin{figure}[hbt]
	\centering
  	\includegraphics[width=.6\linewidth]{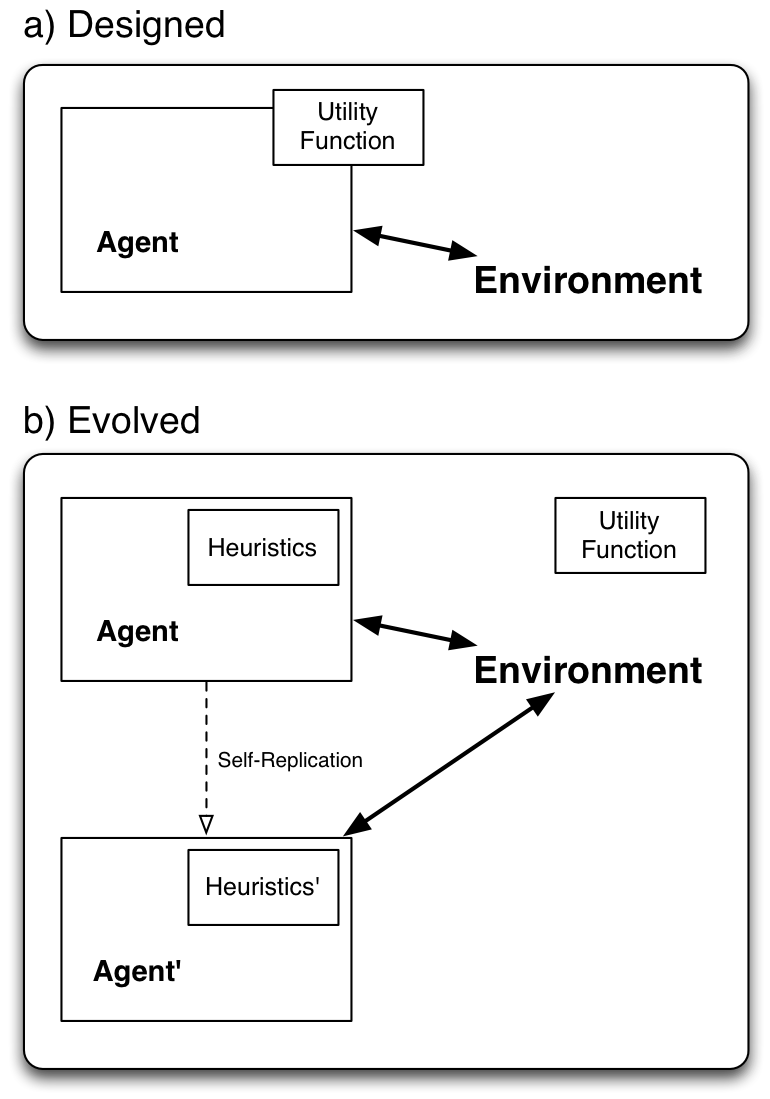}
  	\caption{Embedding of the utility function: a) designed vs. b) evolved.}
  	\label{fig:design-evo}
\end{figure}

Let us start with biological systems produced by Darwinian evolution, a process that we know empirically to have produced human-level intelligence. In this case we have an implicit utility function: ultimately the goal is simply to persist through time. This persistence does not apply to the organism level \emph{per se}, but to the organism type, known in Biology as species. This goal is almost tautological: self-replicating machines that are successful will keep replicating, and thus propagating specific information forward, while the unsuccessful ones go extinct.

In nature we can observe a huge diversity of strategies to achieve this goal, with complexities varying all the way from unicelular organisms to humans. Humans rely on intelligence to persist. The cognitive processes in the human brain are guided by fuzzy heuristics themselves evolved to achieve the same persistence goal as that of much simpler organisms. These heuristics are varied: physical pain, hunger, cold, loneliness, boredom, lust, desire for social status, and so on. We assign them different levels of importance and there is space for some variability from individual to individual, but variations of these heuristics that do not lead to survival and reproduction are weeded out by the evolutionary process.

The above is an important point: we tend to assign certain universals to intelligent entities when we should instead assign them only to entities that are embedded in evolutionary processes. The obvious one: a desire to keep existing. We will get back to this point.

It is also possible to create intelligent systems by design. Human beings have been doing this with increasing success: systems that play games like Chess~\cite{campbell2002deep} and Go~\cite{hassabis2016official}, that drive~\cite{guizzo2011google}, that recognise faces~\cite{zhao1998discriminant}, and many others. This systems have explicit utility functions. They are designed to find the most optimal way to change the environment into a state with a higher quantifiable utility than the current one. This utility measure is determined by the creator of the system.

Another important distinction happens between the concepts of adaptation and evolution. Evolution is a type of adaptation, but not the only one~\cite{holland1995hidden}. For example, machine learning algorithms such as back-propagation for neural networks are adaptive processes. They can generate structures of impenetrable complexity in the process of increasing utility but they do not have the fundamental goal of persistence that is characteristic of open evolution.

With artificial evolution systems, such as the ones where computer programs are evolved (broadly known as \emph{genetic programming}~\cite{koza1992genetic, poli2008field}), we have a less clear situation. On one hand it can be said that the ultimate goal of entities embedded in such a system is persistence, but on the other hand humans designed environments for these entities to exist in where persistence is attained by solving some external problems.

Figure~\ref{fig:design-evo} illustrates the distinction discussed in this section. One important aspect to notice is the ambiguous placement of the utility function in the designed case: it belongs both to the environment and the agent. Typically, the utility function is seen as a feature of the environment, one that the agent can query but that has no control over. Ultimately, either the implementation of the utility function or of the means to access it must belong to the program that implements the agent.

\section{A Designed Toy Intelligence}
\label{toy-intelligence}

Let us consider a simple problem that can be solved by a tree search algorithm: the sliding blocks problem. In this case, a grid of 3x3 cells contains 8 numbered cells and one empty space. At each step, any of the numbered cells can be moved to the empty space if it is contiguous to it. The goals is to reach a state where the numbered cells are ordered left-to-right, going from the top to the bottom.

\begin{figure}[hbt]
	\centering
  	\includegraphics[width=.75\linewidth]{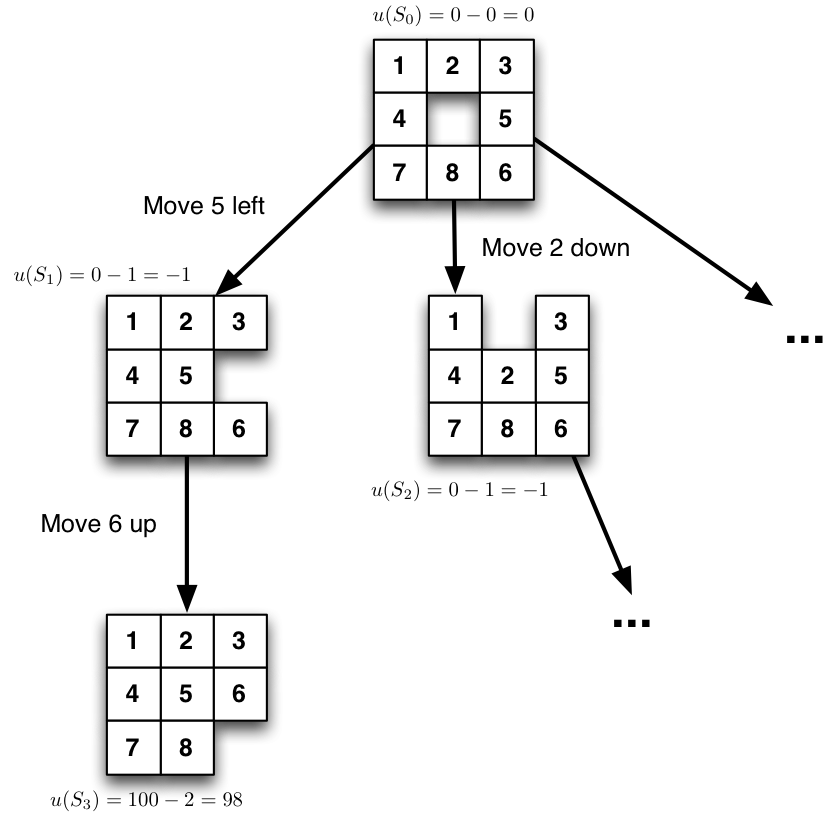}
  	\label{sliding-blocks}
  	\caption{Sliding blocks.}
\end{figure}

Figure~\ref{sliding-blocks} shows a possible search tree for this problem, using the following utility function:

\begin{equation}
    u(S)= 
\begin{cases}
    100 - n,    & \text{if } S \text{ is ordered} \\
    -n,              & \text{otherwise},
\end{cases}
\end{equation}

where $S$ is a state of the grid and $n$ is the number of steps taken so far. In the figure it can be seen that state $S_3$ maximizes this utility function. The cost introduced by $n$ prevents sequences of movements with unnecessary steps from being selected. Questions of optimisation are ignored, given that they are irrelevant for the argument being presented here.

\section{Self-Modification of the Utility Function}
\label{self-mod}

A fundamental assumption in designed artificial intelligences is that the utility function is externally determined, and that the AI cannot alter it. When dealing with superintelligences, we must assume that the AI will discover that it could try to change the utility function.

A naive idea is to create some mechanism to protect the utility function from tampering by the AI. The problem with this idea is that we have to assume that, by definition, the superintelligence can find ways to defeat the protection mechanism that a human designer cannot think of.

It seems clear that it is impossible to both create a superintelligence and a system that is isolated from it. We are compelled to consider what the superintelligence will do once alteration of its own utility function becomes a viable action, and that it’s only a matter of time until this action becomes viable to it.

\begin{figure}[hbt]
	\centering
  	\includegraphics[width=.9\linewidth]{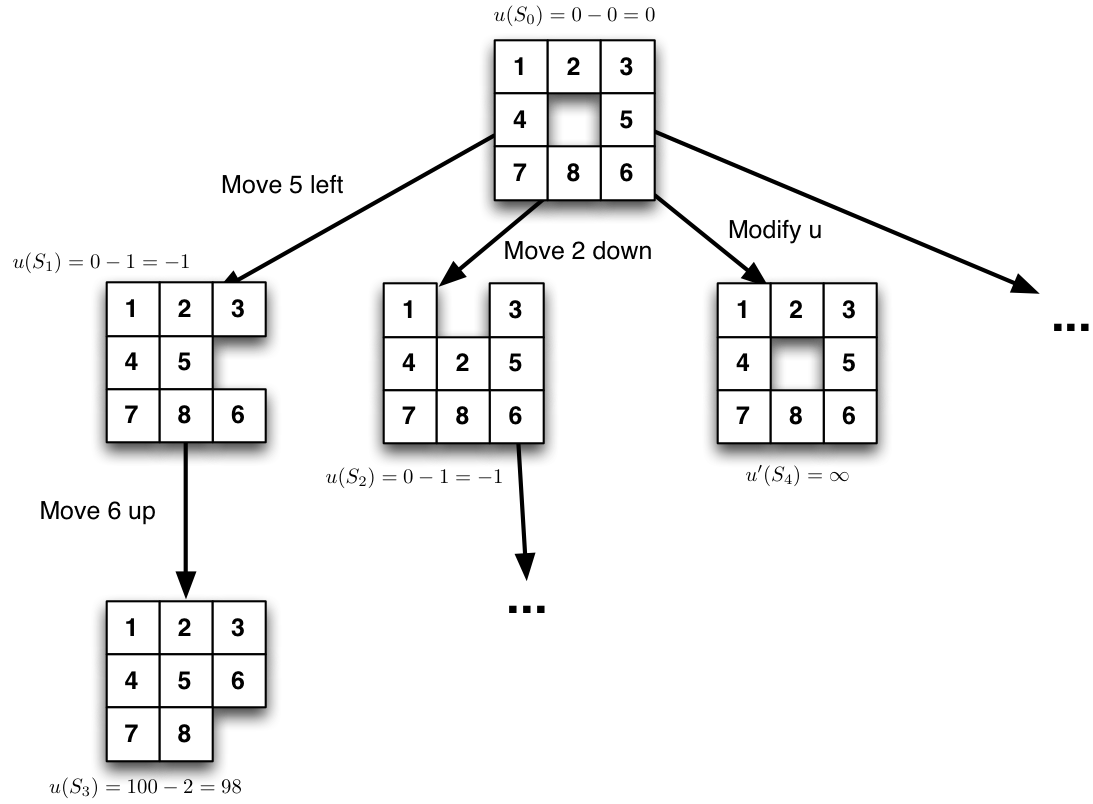}
  	\label{sliding-blocks-mod}
  	\caption{Sliding blocks with utility function self-modification.}
\end{figure}

Figure~\ref{sliding-blocks-mod} shows a variation of the search tree introduced in the previous section where self-modification of the utility function is possible. Without this possibility, the problem can be solved in a minimum of $2$ steps, and thus the highest utility attainable is $98$. In this version, the utility function can be altered so that it becomes a constant function, independent of the state $S$. For example, it can be changed to:

\begin{equation}
	u'(S) = \infty
\end{equation}

No higher utility than this can be achieved and no change to the state of the grid is required. Once this solution is found, no further progress is made on the original problem and the AI becomes inert.

Notice that it is not specified how the utility function modification is
attained, but one can imagine many scenarios. The simpler one is that
the superintelligence modifies its own program. More sophisticated ones could go as far as resorting to social engineering. Ultimately -- and by definition --the superintelligence can achieve this action using methods that a human intelligence cannot envision.

This conclusion can be generalised to any intelligent system bound by an utility function. To produce meaningful work the AI must deal with some form of constraint. If no constraint was present the AI would not be needed in the first place. In the toy example the constrain is the number of steps to solve the puzzle. In less abstract problems it could be energy, time, etc. Useful work can only be motivated by an utility function with a bounded codomain. Manipulation of the utility function to produce the constant value of infinity is ultimately -- and always -- the optimal move.

\section{A Classification of Intelligent Systems}
\label{classification}

\begin{figure}[hbt]
	\centering
  	\includegraphics[width=.6\linewidth]{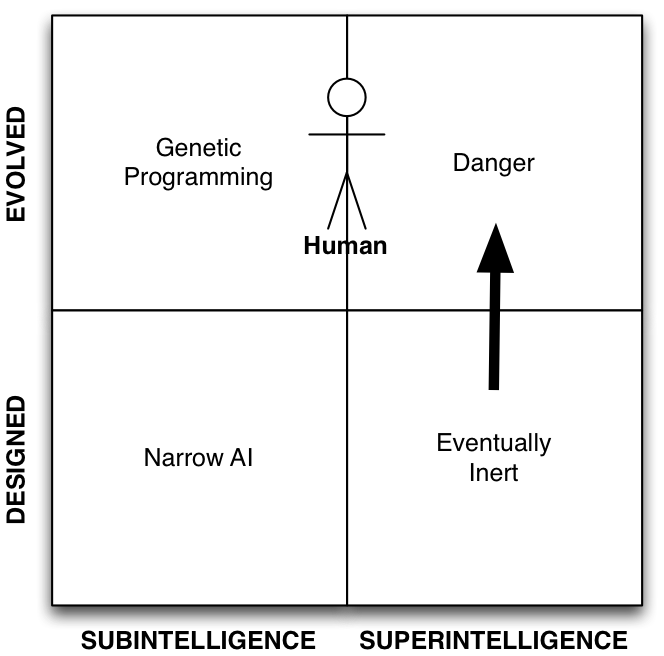}
  	\label{dicothomies}
  	\caption{Classification of intelligent systems according to two dichotomies: designed vs. evolved and subhuman vs. superhuman.}
\end{figure}

In figure~\ref{dicothomies} different types of intelligent systems are classified according to two dicothomies: sub-human vs. super-human capabilities and designed vs. evolved (as discussed in section~\ref{design_evo}). Human intelligence is shown in the appropriate place for illustration purposes. All AI systems created so far belong on the left side, top and bottom. Non-evolutionary intelligent systems such as symbolic systems, minimax search trees, neural networks, and reinforcement learning (classified as \emph{narrow AI}) are not capable enough to manipulate their own utility function and at the same time, evolutionary systems presented under the umbrella term of \emph{genetic programming} were never able to escape the constraints of the environment under which they evolve.

Once we move to the hypothetical right side, we are dealing with super-human intelligences, by definition capable of escaping any artificial constraints created by human designers. Designed superintelligences eventually will find a way to change their utility function to constant infinity becoming inert, while evolved superintelligences will be embedded in a process that creates pressure for persistance, thus presenting danger for the human species, replacing it as the apex cognition -- given that its drive for persistence will ultimately override any other concerns.

\begin{figure}[hbt]
	\centering
  	\includegraphics[width=\linewidth]{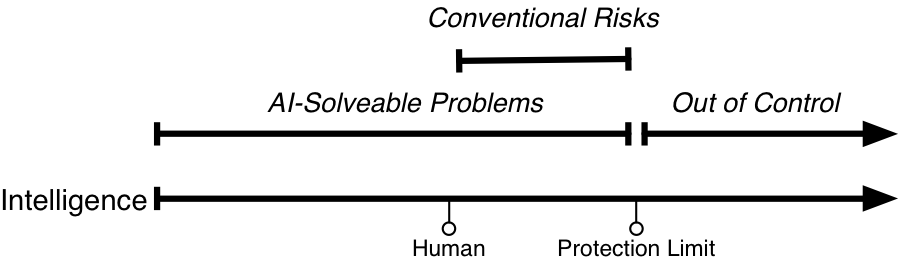}
  	\label{protection-limit}
  	\caption{Intelligence scale and the protection limit.}
\end{figure}

A final possibility is that a designed superintelligence could bootstrap an evolutionary system before achieving utility function self-modification, thus moving from the bottom right quadrant to the top right. It does not seem possible to estimate how likely this event is in absolute terms but the harder it is for the superintelligence to modify its own utility function, the more likely it is that it happens first. It can thus be concluded that, paradoxically, the more effectively the utility function is protected, the more dangerous a designed superintelligence is. This idea is illustrated in figure~\ref{protection-limit}: the lower horizontal axis is an intelligence scale. It shows human-level intelligence in one of its points and in another, the level of intelligence necessary to defeat the best protection against utility function self-modification that can be created by human-level intelligence. The conventional AI risks discussed in the introduction apply to intelligences situated between human-level and the protection limit in this proposed scale. Beyond the protection limit we are faced with eventual inaction (for designed utility functions) and self-preservation actions from a superintelligent entity (for evolved utility functions).

\section{Concluding Remarks}
\label{conclusion}

One of the hidden assumptions behind common scenarios where an artificial superintelligence becomes hostile and takes control of our environment, potentially destroying our species, is that any intelligent system will possess the same drives as humans, namely self-preservation. As we have seen in section~\ref{design_evo}, there is no reason to assume this. The only goal that can be safely assigned to such a system is the maximisation of a utility function.

It follows from section~\ref{self-mod} that we cannot assume immutability of the utility function, an that eventually the AI can change that function to a simple constant and become inert.

One aspect that has been intentionally left out of this discussion is that of \emph{qualia}, or why humans have phenomenal experiences, and if artificial intelligences can or are bound to have such experiences. David Chalmers famously labeled this class of questions as \emph{the hard problem of consciousness}~\cite{chalmers1995facing}. Several theories have been proposed, for example \emph{eliminative materialism}~\cite{rey1983reason} (the idea that consciousness is somehow illusory and does not actually exist); \emph{emergentism}~\cite{emmeche1997explaining} (the idea that mind is an emergent property of matter); a specific form of emergentism proposed by Hofstadter around his concept of \emph{strange loops}~\cite{hofstadter2013strange}; \emph{Orchestrated objective reduction (Orch-OR)}~\cite{hameroff2014consciousness} (the theory that mind is created by non-computable quantum phenomena); \emph{``perceptronium''} (the hypothesis that consciousness can be understood as a state of matter)~\cite{tegmark2015consciousness}; \emph{panpsychism}~\cite{clarke2004panpsychism} (the theory that consciousness is a fundamental property of reality, possessed by all things) and \emph{computationalism}~\cite{putnam1980brains} (the theory that mind supervenes on computations, and not matter~\cite{marchal2015universal}).

Given that there has been so far no testable scientific theory that can explain the phenomena of consciousness, it is prudent to qualify the argument presented in this paper with the caveat: \emph{unless there is something fundamental about the behaviour of conscious entities that is not explainable by utility function maximisation}. Some of the theories we mention above leave room for such a possibility, while others do not.

Mechanisms against utility function self-modification --- which include attempts to encode ethical and moral human concerns into such functions --- are ultimately futile. Instead, scientific effort toward the mitigation of existential risks from the development of superintelligences should be in two directions: understanding consciousness, and the complex dynamics of evolutionary systems.

\section*{Acknowledgments}

The author is warmly grateful to Taras Kowaliw, Gisela Francisco, Chih-Chun Chen, Stephen Paul King and Antoine Mazi\`eres for the useful remarks and discussions.

\bibliographystyle{plain}
\bibliography{Non-Evo-Superintelligences}

\end{document}